UDC 621.391

# Information quantity in a digital image

***M.V. Kharinov***

***St. Petersburg Institute for Informatics and Automation,***
***Russian Academy of Sciences,***
***14-aya Liniya 39, St. Petersburg, 199178, Russia***
***e-mail: khar@iias.spb.su***

The paper is devoted to the problem of integer-valued estimating of information quantity in a pixel of digital image. The definition of an integer estimation of information quantity based on constructing of the certain binary hierarchy of pixel clusters is proposed. The methods for constructing hierarchies of clusters and generating of hierarchical sequences of image approximations that minimally differ from the image by a standard deviation are developed. Experimental results on integer-valued estimation of information quantity are compared with the results obtained by utilizing of the classical formulas.

**Introduction**

For computer calculations a set of pixels of a digital image is divided into clusters of pixels, which either constitute connected segments, either represent the sets of none-adjacent segments, or any other subsets of pixels, and the image is described in terms of appropriate structural elements. Usually it is assumed that the same pixels contain equal information quantities, estimated from the classical formulas [1]. In accordance with the estimation by R. Hartley the information quantity $q_i$ in the $i$-th pixel of the image is the same for all pixels and is expressed in bits as a binary logarithm of the number $g$ of intensity levels:

$$q_i = \log_2 g \, , \tag{1}$$

and in accordance with the estimation by C. Shannon the information quantity $q_i$ in the $i$-th pixel is calculated as a binary logarithm of pixel probability $p_i$, taken with the opposite sign:

$$q_i = -\log_2 p_i \, . \tag{2}$$

For any pixel subset, particularly a set of pixels of the image, the informa-tion quantity $Q$ is calculated by adding $q_i$ :

$$Q = \sum_{i=1}^{N} q_i \, , \qquad\qquad (3)$$

where $N$ is the number of pixels in the image.

So the integral information quantity $Q$ is expressed as $Q = N \log_2 g$ for estimation by R. Hartley (1). For estimation (2) by C. Shannon, $Q$ has a form

$$Q = -\sum_{i=1}^{N} \log_2 p_i \equiv -N \sum_{i=1}^{g} p_i \log_2 p_i \, , \text{ where in the first formula the summation}$$

is carried out over the pixels, and in the second formula the summation is carried out over the intensity levels.

In both classical estimations (1) и (2) the real values are obtained, that violates interpretation of bit as the smallest unit of information quantity. In our steganographic model of an image as a storage medium [2, 3] this difficulty of interpretation does not arise, since, according to the model, the digital image produces its own "virtual" memory that provides data storing and extracting like a conventional computer memory. In steganography task the information quantity is defined as embedding capacity or the maximum volume of embedded data. At that the information quantity turns out to be an integer. Currently, the model is generalized to solve the problem of optimal image segmentation [4, 5]. The development of the model involves the generalization and refinement of integer-valued estimation of information quantity. The aim of the paper is a definition of integer information quantity in a pixel of digital image and its comparison with the classical estimations.

## 1. Integer estimation of information quantity

Let the clusters of identical pixels are treated as indivisible and referred to as *uniform* clusters.

Let us construct a binary hierarchy of pixel clusters wherein for each non-uniform cluster the division into two sub-clusters with different average intensities is defined. Then the definition of an integer information quantity in a pixel of the image is formulated as follows.

**Definition.** The number of bits of information quantity in a pixel of the image coincides with the number of non-uniform clusters that contains the given pixel.

In the ordinary case of uniform clusters they coincide with the whole sets of pixels of a particular intensity. At that, if the clusters are divided into a hie-rarchy of nested clusters consisting of the same number of uniform clusters, the above definition is expressed by R. Hartley estimation (1). If the hierarchy is so that each cluster is divided into sub-clusters with the same number of pixels, it is expressed by C. Shannon estimation (2).

The meaning of the definition consists in direct counting of binary values of information accompanied with division of cluster into two non-uniform sub-clusters. These binary values can be sequentially read from the pixels of nested sub-clusters depending on their higher or lower average intensity.

## 2. Isomorphic invariant image representation

A binary hierarchy of pixel clusters is perceptually determined by means of so called «compact» representation $Hu$ of the image $u$. Compact representation $Hu$ is such an image in which a set of clusters is packed in a hierarchical sequence of image partitions with minimal repetitions of only uniform clusters.

In our interpretation image information is treated as certain codes, which are extracted from the image. These codes constitute a compact representation of the image information $Hu$ according to the rules of the pseudo-ternary number system [2]. In pseudo- ternary number system, the non-negative integers are expanded in powers of 2 , as in conventional binary number system, but with coefficients 0, 1 and 2 , as in ternary number system . Extracted bits are encoded by values 0 and 2, and the neutral value 1 encodes extracting of zero quantity of information from a pixel of indivisible cluster.

So, the compact representation specifies the hierarchy of clusters encoded in the pixels of $Hu$, which in an iterative top-down mode of splitting of clusters are constructed as follows.

At the first iteration all pixels of the image are assigned to the same cluster, and all pixels of $Hu$ are zeroed.

At the next iteration, each non-uniform cluster of image pixels is split into two sub-clusters, and the pixel values of the invariant representation $Hu$ are doubled. If the given cluster of image pixels is uniform, then all pixels of corresponding cluster of invariant representation $Hu$ are increased by 1. If the given cluster of image pixels is non-uniform, then the pixels of the invariant representation $Hu$, corresponding to the pixels of the image that contained in the nested cluster with a higher average intensity, are increased by 2. Transformation of the remaining pixels of the compact representation $Hu$, corresponding to the pixels of the image from the nested cluster with a lower average intensity, at given iteration is limited to the previous doubling. Iterations of constructing of compact representation $Hu$ are repeated. The procedure is completed when the set of image pixels is decomposed into uniform clusters.

Pseudo-ternary number system for each pixel of compact representation enables to reproduce the values obtained at any iteration. To obtain the sequence of these values the simple arithmetical transformation is iteratively used, consisting in that odd values are entirely divided by 2, and even values are entirely divided by 4 and doubling then [2]. In this case an integer information quantity in given pixel of the image coincides with the number of even values in the discussed sequence of reducing values of pixel of compact representation $Hu$. Thus, if $Hu$ is known, then the integer estimation of information quantity is trivial and doesn't require to resume the image analysis. Interpretation of transformations of pixel values by means of arithmetic operations with powers of 2 as shifts of codes in virtual memory gives grounds to treat $Hu$ as an image representation recorded in the digital memory, which is generated by the image itself and is similar to a conventional computer memory [2, 3].

The compact image representation $Hu$ is also called "invariant", since it remains unchanged under linear transformation of pixel intensities for certain algorithms of splitting of non-uniform clusters into a pairs of nested subclusters, for example, by conventional Otsu thresholding [6].

The transformation $H$ of image $u$ into invariant representation $Hu$ commutes with its conversion into negative, as well as with image scaling by duplicating pixels. At the same time, the mentioned conversions don't influence on information quantity in a pixel of an image.

A characteristic property of invariant image representation at all iterations of constructing is that it is obtained by *isotone* (i.e., order-preserving) converting of average intensities within the clusters of image pixels into integer pixel values of invariant representation $Hu$ without changing the order of intensity levels. In other words, an invariant representation $Hu$ is *isomorphic* to piecewise constant approximation of the image with the average intensities, wherein the pixel values are replaced by the averages within the clusters.

### 3. Quasioptimal image approximations

According to the proposed general definition, a particular integer estimation of information quantity essentially depends on a hierarchy of clusters, which can be defined by the algorithm of non-uniform cluster splitting into pair of sub-clusters, as well as by the sequence of piecewise constant approximations of the image, wherein cluster repetitions exist. The most widely used criterion of quality of approximation is the standard deviation $\sigma$ of the approximation from the image or the total square error $E = N\sigma^2$ [7]. Usually, the approximations that minimally differ from an image in $E$ or $\sigma$ for se-quential cluster numbers are considered as the best. In general, the sequence of optimal image approximations is not hierarchical and therefore insufficiently convenient for calculations [4]. Nevertheless, in image processing tasks it appears possible to majorize nonhierarchical sequence of optimal approximations by hierarchical sequence of quasioptimal approximations [8].

In the simplest case, the sequence of quasioptimal image approximations is obtained by splitting of the non-uniform clusters according to conventional histogram Otsu method [6], wherein the threshold intensity value is found from the condition of maximum decrease of the total squared error $E$. However, Otsu method [6] has no obvious generalization to the case of color and multispectral images. In perspective of multidimensional generalization it seems more promising to define the cluster splitting as backward last step of iterative merging of clusters [4, 9, 10].

To avoid analysis of cluster repetitions the computation of hierarchical sequence of quasioptimal image approximations, containing 1, 2, 3, ... clusters of pixels, is performed in two stages. At the first stage a compact invariant representation $Hu$, which specifies the sequence of partitions of the image pixels into 1, 2, 4, 8 ... clusters, is calculated. At the second stage, a compact representation is expanded into a sequence of approximations with successively increasing numbers of clusters, providing the maximal decrease of the total squared error $E$ or standard deviation $\sigma$.

The results of calculations are presented either as ordinary piecewise constant image approximations with 1, 2, 3, 4 ... average intensities, either as corresponding invariant representations that describe the piecewise constant approximations of the averaged pixels in a similar manner as $Hu$ describes the original image.

Distinguishing feature of quasioptimal image approximations with 1, 2, 3, 4 ... average intensities is that the corresponding sequence $E_1, E_2, E_3, ...$ of values of the total squared error $E$ is convex [10]:

$$E_i \leq \frac{E_{i-1} + E_{i+1}}{2}, \quad i = 2, 3, ..., g - 1. \qquad (4)$$

Convexity property (4) holds also for a compact sequence of quasioptimal approximations of 1, 2, 4, 8 ... clusters.

Thus, quasioptimal approximations preserve the property (4) of optimal approximations. In contrast to the optimal approximations, quasioptimal approximations are convertible into isomorphic invariant representations.

## 4. Experimental results

To illustrate the content of the paper we present the results of computations obtained for a standard image "Lena" (Fig. 1).

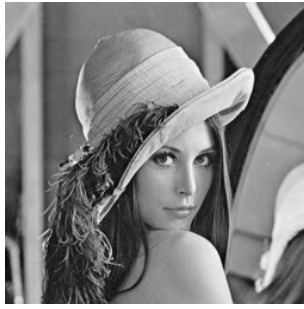

Fig. 1. Standard image "Lena".

Fig. 2 allows evaluating the quality of quasioptimal approximations compared with optimal [8].

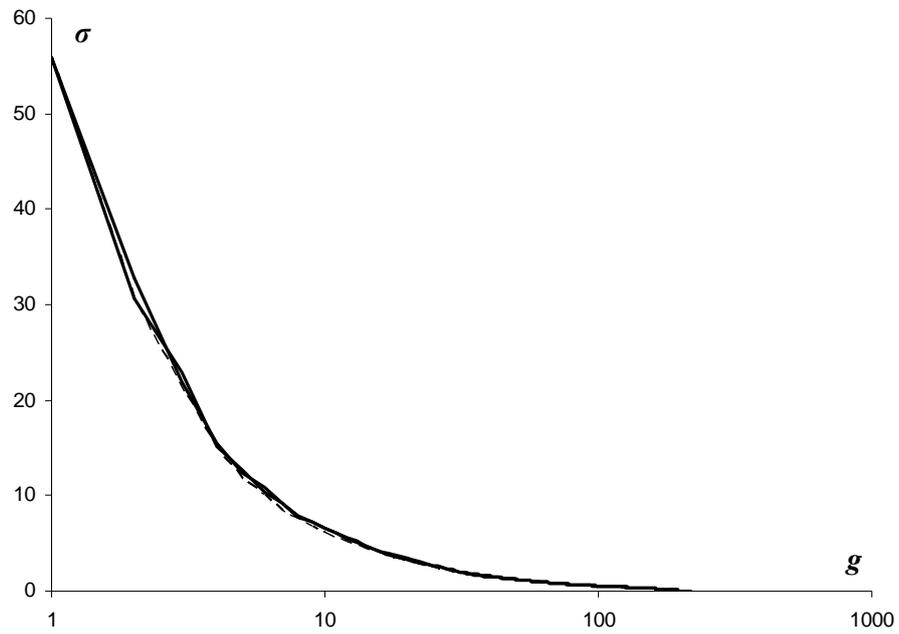

Fig. 2. A standard deviation $\sigma$ of optimal and quasioptimal approximations from the image depending on the number $g$ of pixel clusters.

Fig. 2 on a logarithmic scale along the abscissa shows the graphs of standard deviation $\sigma$ of the number of clusters $g$ for optimal approximations (bottom dashed curve) as well as the approximations, calculated by the iterative Otsu thresholding, and approximations obtained by iteratively merging clusters (two upper solid interlaced curves).

As seen in Fig. 2, the curves practically coincide with each other that illustrates the possibility of approximating of non-hierarchical optimal approximation sequence by hierarchical sequence of quasioptimal approximations, which are much simpler to calculate, store and analyze when computing.

Fig. 3, for two, three and four levels of average intensity, demonstrates the optimal approximations and quasioptimal image approximations.

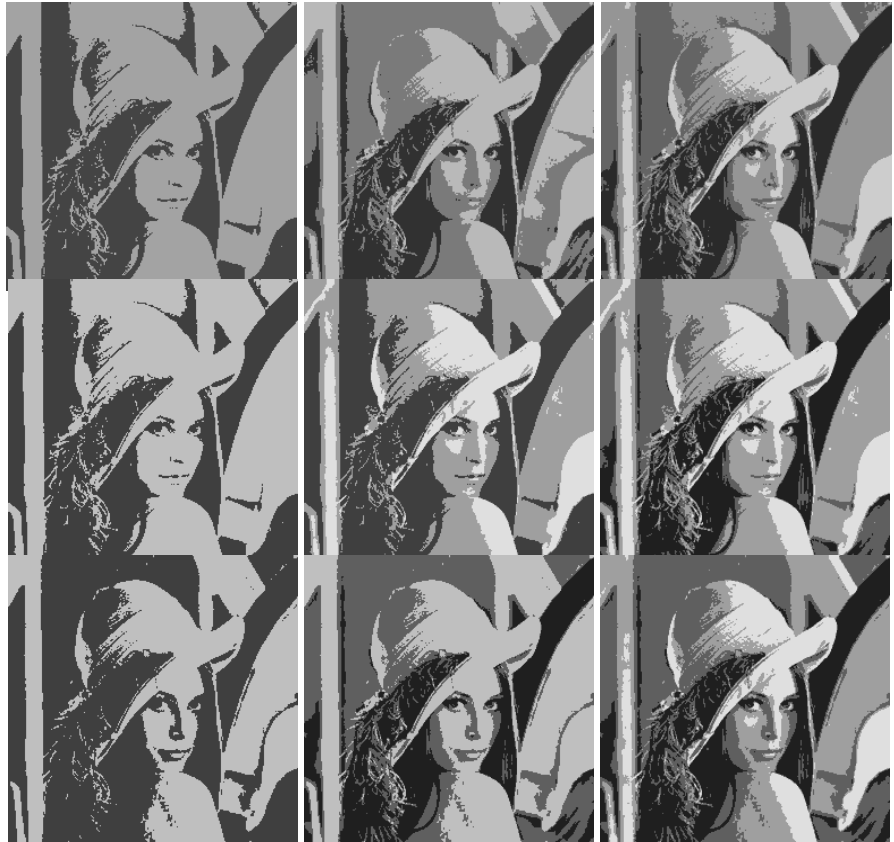

Fig. 3. Optimal and quasioptimal piecewise constant approximations.

Optimal approximations are placed in the top row, and quasioptimal image approximations are listed in two bottom rows. The latter are shown in the invariant mode of representation, normalized for working intensity range.

Approximations obtained by iterative Otsu thresholding are shown in the middle row. Approximation based on the cluster splitting by means of reversing of last merging of clusters [4, 9, 10] are shown in the bottom row. Visually, the quasioptimal approximations reproduce the image no worse than optimal approximations. At the same time, owing to the property of isomorphism, the invariant representation has a little effect on perception, but in pattern recognition can significantly improve the robustness of image segmentation results.

The *information quantity in the image approximation* is estimated as the information quantity in the image, wherein pixel values are replaced by intensities, averaged within each of $g$ clusters constituting the given approximation, so that these clusters are formally treated as indivisible.

In this case, the information of the image is divided into information encoded in $g$ clusters as independent images and the information encoded in the image by means of a clusters, considered as an elements of the image. Then, the integral information quantity $Q$ in an image is decomposed into the sum of the information quantity $Q_0$ in the approximation of an image and information quantities $Q_i$ in $g$ clusters, considered as separate images:

$$Q = Q_0 + \sum_{i=1}^{g} Q_i \,, \tag{5}$$

Property (5) of information quantity can be regarded as strengthened monotony property, stating that the information quantity of the whole image is not less than the total information quantities in its parts: $Q \geq \sum_{i=1}^{g} Q_i$ .

The graphs in Fig. 4 illustrate for integer-valued and classical estimations the behavior of total information quantity in the image approximations depending on the number of clusters $g$ .

Integral quantity of information $Q_0$ is measured by the percentage of the image volume in a computer memory. Dashed curves in Fig. 4 describe estimations by C. Shannon. Estimations by R. Hartley are marked with gray curves. Our integer-valued estimation is shown by black solid lines.

The left graph in Fig. 4, constructed from the data of [3], meets the hierarchy of clusters designed for steganographic embedding. When constructing, the division into sub-cluster is carried out by minimizing the absolute value of the difference between the numbers of pixels in nested sub-clusters.

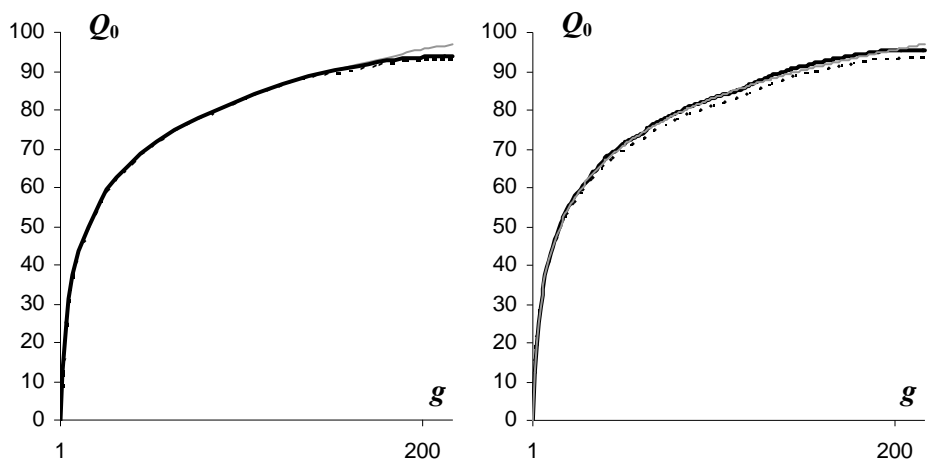

Рис. 4. A comparison of the integer estimation of information quantity with the classical estimations.

In this case an integer estimation of information quantity has intermediate values between the classical estimations and with increasing number of clusters approaches to estimation by C. Shannon. As shown in [3], not only the integral value of integer estimation of information quantity are consistent with estimation by C. Shannon, but also its distribution over the pixels of the image. Thus, if the total square error $E$ is not taken into account when constructing the image approximations, then we get an integer analogue of Shannon's estimation.

Graph on the right of Fig. 4 illustrates the behavior of information quantity estimations for the approximations constructed by minimizing of the total squared error in the algorithm of cluster splitting by reversing of last step of iterative cluster merging [4, 9, 10]. In this case, the curve for integer-valued estimation is somewhat higher than the curve for the Shannon's estimation and is intertwined with the curve for Hartley's estimation. So, while minimizing the total squared error $E$ we come to an integer-valued analogue of Hartley's estimation. It should be added that the algorithm of iterative Otsu's thresholding leads to the similar results.

In general, basing on the results of the experiments, we can conclude that the proposed integer estimation agrees with the classical estimations of information quantity.

## Conclusion

In this paper we have tried to develop and generalize the concept of integer-valued quantity of information, avoiding the details of previously developed and patented steganographic model of image with virtual memory [2, 3], in particular, the details of data embedding in the image (RF patents №2006119146 and №2006119273). Focusing on extracting information, we have formulated the definition of an integer information quantity and checked it for plausibility by comparing the results with those obtained by the classical formulas.

At the current stage of the study, the proposed integer estimation of information quantity in the image is practically important for steganography applications. It may seem that, in another image processing tasks, particularly in the task of image segmentation by means of optimal approximations, the application of the integer estimation of information quantity and related notions is limited by histogram processing methods. However, it is not quite so. Effective majorizing of optimal image approximations by means of quasioptimal approximations enables one to reduce the optimal image approximating by connected segments to a simple reduction in the number of segments that constitute pixel clusters of quasioptimal approximations.

Methods of constructing of quasioptimal image approximations with a limited number of connected segments and study of the peculiarities of corresponding integer-valued estimating of the information quantity determine the direction of future research.

## References

1. R.M. Yusupov «*Theoretical Foundations of Applied Cybernetics. Issue 1. Elements of Information Theory*», 1973. A.F.Mozhayskij VIKA, Leningrad, Russia, 110 p. [in Russian].

2. M.V. Kharinov «*Storage and Adaptive Processing of Digital Image Information*», 2006. St. Petersburg University Press, St. Petersburg, Russia, 138 p. [in Russian]. ISBN: 5-288-04209-8

3. M.V. Kharinov, V.P. Zabolotsky "Steganographic document protection based on a model image information storing", Information and Communication. 2010, No 1, Moscow, Russia, pp. 77-81. ISSN 2078–8320. [in Russian].

4. M.V. Kharinov «A model for localization of objects in digital images», Bulletin of the Buryat State University. 2013. Number 9, Ulan-Ude, Russia, pp. 182-189. ISSN 1994-0866 [in Russian].

5. M.V. Kharinov «Stable image segmentation», Bulletin of the Buryat State University. 2012. No 9, Ulan-Ude, Russia, pp. 64-69. ISSN 1994-0866 [in Russian].

6. N. Otsu «A Threshold Selection Method from Gray-Level Histograms», IEEE Transactions on systems, MAN, and CYBERNETICS. January 1979. Vol. SMC-9, Number 1, pp. 62-66.

7. S.A. Aivazyan, V.M. Buchstaber, I.S. Eniukov, L.D. Meshalkin «*Applied Statistics: Classification and Reduction of Dimension*», 1989. Finance and Statistics, Moscow, Russia, 607 p. [in Russian].

8. M. V. Kharinov Image Segmentation Method by Merging and Correction of Sets of Pixels // Pattern Recognition and Image Analysis: Advances in Mathematical Theory and Applications / 2013. Pleiades Publishing, Ltd. Vol. 23. No 3, pp. 393-401, ISSN 1054-6618.

9. A.Z. Arifin and A. Asano «Image segmentation by histogram thresholding using hierarchical cluster analysis», Pattern Recogn. Letters, 2006. Number 27 (13), 1515-1521.

10. M.V. Kharinov Image segmentation by optimal and hierarchical piecewise constant approximations // 11th International Conference on Pattern Recognition and Image Analysis: New Information Technologies (PRIA−11−2013). Proc. of the 11-th. Int. Conf. Samara: IPSI RAS, September 23-28, 2013. Vol. 1. pp. 213-216.

*Kharinov Mikhail Vyacheslvovich*, candidate of technical sciences, senior researcher of Laboratory of Applied Informatics of St. Petersburg Institute for Informatics and Automation of RAS.